\documentclass[letterpaper, 10 pt, conference]{ieeeconf}  % Comment this line out if you need a4paper

\IEEEoverridecommandlockouts                              % This command is only needed if 
\usepackage{cite}
\usepackage{amsmath,amssymb,amsfonts}

\usepackage{graphicx}
\usepackage{textcomp}
\usepackage{xcolor}
\usepackage{hyperref}
\usepackage{multirow}
\usepackage{hhline}
\usepackage{subcaption}
\usepackage{algorithm}
\usepackage{algorithmic}
\usepackage{array}
\usepackage{stfloats}
\usepackage{url}
\usepackage{enumerate}
\usepackage{float}
\usepackage[normalem]{ulem}
\usepackage{yhmath}
\usepackage[export]{adjustbox}
\usepackage{diagbox}
\usepackage{kotex}
\usepackage{verbatim}
\usepackage[margin=0.75in]{geometry}
\renewcommand{\baselinestretch}{0.925}
\usepackage{amsmath}

\newcommand{\rvauthcommands}[4]{%
\expandafter\newcommand\csname #1rev\endcsname{#4} % rev number
\expandafter\newcommand\csname #1t\endcsname[2][0]{\ifnum##1>\csname #1rev\endcsname\textcolor{#3}{\sout{##2}}\fi} % remove without author indication
\expandafter\newcommand\csname #1r\endcsname[3][0]{\ifnum##1>\csname #1rev\endcsname\textcolor{#3}{\sout{##2}##3}\else##3\fi} % replace without author indication
\expandafter\newcommand\csname #1c\endcsname[2][0]{\ifnum##1>\csname #1rev\endcsname\textcolor{#3}{/\uline{##2}/}\fi} % comment without author indication
\expandafter\newcommand\csname #1\endcsname[2][0]{\ifnum##1>\csname #1rev\endcsname\textcolor{#3}{##2}\else##2\fi} % add without notice
}% define custom rv commands
\title{\LARGE \bf
Semi-Supervised Imitation Learning with \\ Mixed Qualities of Demonstrations for Autonomous Driving 
}

\author{ Gunmin Lee$^{1}$, Wooseok Oh$^{1}$, Seungyoun Shin$^{2}$, Dohyeong Kim$^{1}$, Jeongwoo Oh$^{1}$, Jaeyeon Jeong$^{1}$, \\
Sungjoon Choi$^{3}$, and Songhwai Oh$^{1}$
\thanks{$^{1}$G. Lee, W. Oh, D. Kim, J. Oh, J. Jeong, and S. Oh are with the Department of Electrical and Computer Engineering and ASRI, Seoul National University, Seoul 08826, Korea (e-mail: gunmin.lee@rllab.snu.ac.kr, wooseok.oh@rllab.snu.ac.kr, dohyeong.kim@rllab.snu.ac.kr, jeongwoo.oh@rllab.snu.ac.kr, jaeyeon.jeong@rllab.snu.ac.kr, songhwai@snu.ac.kr).}%
\thanks{$^{2}$S. Shin is with the School of Computer Science and Engineering
Dongguk University, Seoul 04620 (e-mail:2018112005@dongguk.edu).}%
\thanks{$^{3}$S. Choi is with the Department of Artificial Intelligence, Korea University, Seoul 02841, Korea (e-mail:sungjoon-choi@korea.ac.kr).}%
}

\begin{document}

\maketitle

%%%%%%%%%%%%%%%%%%%%%%%%%%%%%%%%%%%%%%%%%%%%%%%%%%%%%%%%%%%%%%%%%%%%%%%%%%%%%%%%
\begin{abstract}

In this paper, we consider the problem of autonomous driving using imitation learning in a semi-supervised manner. In particular, both labeled and unlabeled demonstrations are leveraged during training by estimating the quality of each unlabeled demonstration. If the provided demonstrations are corrupted and have a low signal-to-noise ratio, the performance of the imitation learning agent can be degraded significantly. To mitigate this problem, we propose a method called semi-supervised imitation learning (SSIL). SSIL first learns how to discriminate and evaluate each state-action pair's reliability in unlabeled demonstrations by assigning higher reliability values to demonstrations similar to labeled expert demonstrations. This reliability value is called leverage. After this discrimination process, both labeled and unlabeled demonstrations with estimated leverage values are utilized while training the policy in a semi-supervised manner. The experimental results demonstrate the validity of the proposed algorithm using unlabeled trajectories with mixed qualities. Moreover, the hardware experiments using an RC car are conducted to show that the proposed method can be applied to real-world applications. 

\end{abstract}

%%%%%%%%%%%%%%%%%%%%%%%%%%%%%%%%%%%%%%%%%%%%%%%%%%%%%%%%%%%%%%%%%%%%%%%%%%%%%%%%
\section{Introduction}

%%%%%%
% Drawbacks of existing imitation learning methods
% 1. collecting expert data is expensive, time consuming task + data corruption
% 2. Using non-expert(negative) demo -> solution (can be seen as data corruption)
% 3. Explain existing methods using negative demo
% 4. Drawbacks of methods using negative demo--> need to be classified as negative
% 5. We will solve this problem!-->by leveraging using uncertainty(confidence)
% In this regard, we alleviate this issue by using negative demonstrations.
% Compared to expert demosntrations, diverse-quality demonstrations can be collected more cheaply, e.g., via crowdsourcing (Mandlekar et al., 2018).
%trajectory label을 한개의 distribution의 요소중하나로 선택하여, action, state, trajectory label로 부터 새로운 distribution설정후, 여기서부터 새로운 action u_t로 다음 state에 대입. --> VILD
%
%our contribution: a method to sort out and leverage unlabeled values, use those unlabeled values to training
%%%%%%

Designing a real-time controller for robots with high-speed motion is a challenging problem. Conventional approaches of designing a rule-based feedback controller \cite{rulebasecite2, rulebasecite1,rulebased1} have shortcomings in covering all the exception cases. Recent approaches include a method that can autonomously cover all the exception cases, such as training the controller via deep reinforcement learning (RL) methods \cite{b14, ppo,  RLbased3, RLbased4,RLbased5}. However, it is challenging to use RL methods for physical robots due to their high sample complexity, safety concerns, and difficulty in designing feasible reward functions \cite{challengerl}. 

To address the above issues, imitation learning (IL)\cite{BC, b1, b22, LeeNIPS, MMIL, infogail} has been a trending topic of designing a controller. IL methods do not need to design a reward function and tend to show better performance due to their simplicity of following an ideal trajectory. In general, the main objective of imitation learning is to find a controller that successfully mimics the given expert demonstrations.

Standard approaches to imitation learning (IL)\cite{b1, b22, LeeNIPS, MMIL, infogail} assume that only the expert data are given to the agent. However, this assumption may not always be valid. One of this assumption's main limitations is that collecting pure expert data is challenging due to the limited budget and time. The data may be noisy, not collected from experts, or contain incomplete demonstrations. This causes the learning quality of imitation learning to degrade. On the other hand, collecting diverse quality data is cheaper (i.e., collecting demonstrations from non-experts or crowdsourcing \cite{crowdsource}). However, it is not straightforward to evaluate and discriminate such data for training due to the variability in quality.  

To resolve this problem, Choi et al.\cite{SungjoonTRO, b4} proposed a method that can utilize demonstrations with mixed qualities by estimating the quality of each data through the lens of correlation. The cornerstone of the method is to model multiple policies with correlated Gaussian processes by assuming that the expert's policy is the dominant pattern in the dataset. This method was extended to neural networks in \cite{b8} to address the scalability issue of Gaussian process regression. Mixed demonstration generative adversarial imitation learning (MixGAIL), proposed in \cite{b30}, is an imitation learning method that can be trained with both good and bad demonstrations. But it is assumed that all demonstrations are labeled.

This paper assumes the imitation learning scenario where a relatively small number of expert demonstrations are given while there are enough unlabeled and perhaps suboptimal demonstrations. This setting alleviates the clean data assumption that most imitation learning papers take and makes it possible to leverage both labeled and unlabeled demonstrations in one unified framework. %Also, this paper does not assume that the expert's policy is the dominant pattern in the given dataset.

In this paper, we introduce a novel imitation learning method named \textit{semi-supervised imitation learning} (SSIL) which can evaluate the quality of unlabeled data using the predictive uncertainty. First, we train a variational autoencoder (VAE) from a small number of demonstrations from experts. From the VAE we trained, we can measure the reconstruction losses on unlabeled demonstrations. Then, we normalize the reconstruction loss and convert them into the confidence value of each state-action pair in each unlabeled demonstration. This confidence value is called the leverage value of the given demonstration. We then provide a state-action-leverage value triplet set to the SSIL framework.

We show the strength of the proposed method by comparing it against MixGAIL \cite{b30} and GAIL \cite{b1}. The experiments are performed in TORCS \cite{b12} and a real-world RC car driving scenario. SSIL shows superior performance compared to GAIL and MixGAIL when only more than 0.61\% of demonstrations are given as labeled expert demonstrations. Moreover, in a real-world experiment with an RC car, SSIL shows maximum score improvements of 19.0\% and 12.1\% over GAIL and MixGAIL, respectively. In addition, SSIL shows a similar standard deviation of the performance score on three different trials with different seeds as MixGAIL, which is 58.1\% of the standard deviation of the performance score of GAIL. Note that to show the robustness of our algorithm, the domain for collecting data and the domain used for testing our algorithm are different.

\section{Background}
%%%%%%%%%%%%%%%%%%%%%%%%%%%%%%%%%%%%%%%%%%%%%%%%%%%%%
%1. MixGAIL
%2. VAE reconstruction loss
%%%%%%%%%%%%%%%%%%%%%%%%%%%%%%%%%%%%%%%%%%%%%%%%%%%%%%

\subsection{Mixed Demonstration Generative Adversarial Imitation Learning}
\label{ssec:MixGAIL}
Generative adversarial imitation learning (GAIL) is a powerful imitation learning method. In this algorithm, the agent imitates the given expert demonstration. 
While GAIL \cite{b1} uses only expert demonstrations, mixed demonstration generative adversarial imitation learning (MixGAIL) \cite{b30} uses demonstrations with poor qualities (negative demonstrations) along with expert demonstrations. %As our proposed algorithm also exploits different qualities of demonstrations, MixGAIL is one of our baseline algorithms.

The objective function of MixGAIL is given as:
\begin{equation}
\small\begin{aligned}
\underset{\pi\in\Pi}{\text{min }}\underset{D}{\text{max}}
& E_\pi[\log(D(s, a))-\eta \mathrm{CoR}(s, s_E, s_N)] \\ 
& + E_{\pi_E}[\log(1-D(s, a))]-\lambda H(\pi),
\end{aligned}
\label{mixgaileq}
\end{equation}
where $\pi_E$ and $\pi$ are stochastic policies of expert trajectory and generator, respectively. $E_\pi$ and $E_{\pi_E}$ are the expected returns of policy $\pi$ and $\pi_E$, respectively. The state is $s\in S$, and action is $a\in A$, where $S$ and $A$ are the state and action spaces. $D:S\times A\mapsto(0,1)$ is a discriminator, and the causal entropy of $\pi$ is represented as $H(\pi)$. $s_E$ is a set of states of expert demonstrations, while $s_N$ is a set of states in negative demonstrations. $\mathrm{CoR}$ is a term called the \textit{constraint reward}. $\eta$ and $\lambda$ control the proportion of $\mathrm{CoR}$ and $H$, respectively.
The constraint reward, $\mathrm{CoR}$, is given as:
\begin{eqnarray}
\small
\begin{aligned}
\mathrm{CoR}(s,s_E,s_N) &= \frac{(1+\frac{d_{s_E}}{\alpha})^{-\frac{\alpha+1}{2}}}{(1+\frac{d_{s_E}}{\alpha})^{-\frac{\alpha+1}{2}}+(1+\frac{d_{s_N}}{\alpha})^{-\frac{\alpha+1}{2}}},\\
d_{s_E}&=\sqrt{\frac{1}{|s_E|} \sum\limits_{s_e\in s_E} (\|s-s_e\|_2)^2},\\ 
d_{s_N}&=\sqrt{\frac{1}{|s_N|} \sum\limits_{s_n\in s_N} (\|s-s_n\|_2)^2},
\end{aligned}
\end{eqnarray}
where $\|\cdot\|_2$ is $l_2$ norm and $\alpha$ is a hyperparameter used to control the sharpness of $\mathrm{CoR}$. $\mathrm{CoR}$ estimate how much the current state is close to expert states or negative states. %If $\mathrm{CoR}$ value is low, the current state is close to negative states, and vice versa.
This value is used as an estimated reward function to the policy network before the discriminator is moderately trained.
%In addition, the authors have shown that a policy function can be learned by alternatively updating the discriminator and the policy since (\ref{mixgaileq}) has a unique saddle point.

%new content from here
\subsection{Using Reconstruction Loss as a Measurement of Uncertainty}
A variational autoencoder (VAE) is an encoder-decoder model where the dimension of the given dataset is reduced by the encoder then reconstructed by the decoder. The objective of VAE is to reconstruct the original data from a deducted dimension form, using the decoder model to generate data similar to original data. The loss function of this algorithm is called the reconstruction loss. 

An et al.\cite{bottleneck} have shown that the variational autoencoder can be used for anomaly detection. When data comes across an information bottleneck between the encoder and decoder, the VAE will learn to lose the least amount of information. If unseen data given to the trained VAE is distinct from the data used for training, the reconstruction loss will be high due to the VAE failing to select essential features for reconstruction. If a decent threshold for reconstruction loss is selected, the reconstruction loss can be used as an indicator for anomaly detection. 
\begin{comment}
The reconstruction loss ($L_{Recon}$), which is generally used as l2 loss, is given as:

\begin{equation}
    L_{Recon} = \frac{1}{N}\sum_{i=0}^{N}\|y_{pred_i}-y_{gt_i}\|_2,
\end{equation}
where $N$ is the number of input data, $y_{pred_i}$ is the predicted output of $i$th data by VAE, and $y_{gt_i}$ is the ground truth of $i$th data.

\end{comment}

\section{Semi-Supervised Imitation Learning}
%%%%%%%%%%%%%%%%%%%%%%%%%%%%%%%%%%%%%%%%%%%%%%%%%%%%%%%%
%1. first, say that we know an expert trajectory&negative trajectory, however small.
%2. We then use uncertainty measures to evaluate each data's confidence. When training, the input/output is the expert trajectory we have. Also, we were given some epistemic uncertainty bias to the loss function.
%3. After giving each unlabeled data its uncertainty value, the uncertainty value is changed into a confidence value
%4. the confidence level is given to the GAIL framework with upgraded CoR value(think about it). The new CoR value also only affects the reward function part of GAIL(discriminator+Cor).
%5. show me the results/
%%%%%%%%%%%%%%%%%%%%%%%%%%%%%%%%%%%%%%%%%%%%%%%%%%%%%%%%%
\label{Proposed}
One of the shortcomings of GAIL \cite{b1} is that, without expert demonstrations, GAIL may not be adequately trained. Hence, the performance of using GAIL is limited by the quality of expert demonstrations. Also, if the quality of demonstrations given to the algorithm is not consistent with the policy, the learning process becomes slower, the trained agent may not perform as well, or worse, the agent may not learn from them.
If we have a set of state-action pairs from expert demonstrations, we can evaluate and give corresponding reliability values to other unlabeled state-action pairs and use them as a sub-optimal dataset.
Since each unlabeled demonstration has a different degree of closeness to expert demonstrations, the reliability value for unlabeled state-action pairs has to be continuous, not binary. Therefore, an algorithm using a continuous reliability value is necessary. We call this estimated reliability value a leverage value in this paper. 

We introduce a novel method called semi-supervised imitation learning (SSIL). This method uses both labeled expert demonstrations and unlabeled demonstrations with unknown qualities. 
Our main contributions are twofold. First, we propose a novel method to evaluate each unlabeled state-action pair based on known expert state-action pairs. Second, we propose a method using continuous leverage values for training an autonomous agent.

\subsection{Evaluating Unlabeled Demonstrations}
\label{ssec:confidence}
We propose an algorithm that evaluates unlabeled state-action pairs. The variational autoencoder (VAE) is used to evaluate the quality of a state by calculating the reconstruction loss of unlabeled data. If the reconstruction loss is higher than a predetermined threshold, the data pair corresponding to the reconstruction loss can be classified as harmful data and vice versa. However, binary classification cannot accurately evaluate the data because the data quality is not binary, expert or non-expert, but continuous. Therefore, evaluating the quality of each data is essential for higher performance, especially for autonomous driving.
We obtain a leverage value by normalizing the reconstruction loss as follows: 

\begin{equation}
\begin{aligned}
    L_{Recon} &= \{r_j| r_j = \|s_{in_j}-s_{out_j}\|_2, , 0 \le j \le \|s_T\|\}, \\
    r_{max} &= \max_{\forall r_j \in L_{Recon}} r_j, r_{min} = \min_{\forall r_j \in L_{Recon}} r_j, \\
    l_{T_j} &= \frac{r_j-r_{min}}{r_{max}-r_{min}}, \\
    l_T &= \{l_{T_j}| \forall j \in \mathbb{N}, 0 \le j \le |s_T| \},
\end{aligned}
\label{eq:l_cal}
\end{equation}
where $L_{Recon}$ is the reconstruction error set, $r_j$ is the reconstruction error of $j$th state, $s_{in_j}$ and $s_{out_j}$ are the input state and result of VAE, respectively, $l_{T_j}$ is the leverage value of the $j$th state, $l_T$ is the set of leverage values, and $|s_T|$ is the number of states in all demonstrations.

While the mixture density network (MDN) \cite{MDN} and GPR \cite{GaussianProcess} use both state and action data for training, VAE only uses state data as its input, which is a loss of information. To compensate for this issue, we also propose to use two consecutive state data as input to VAE, which is called windowed VAE (windowVAE). By encoding two consecutive state data, the VAE can be seen as encoding the state data and the environment-action interaction. This encoding method allows VAE to encode not only the state data but also the effect of action. The results in Table~\ref{fig:leveragecomparison} shows that the leverage values are evaluated as we intended.

\subsection{Leverage Constraint Reward}
The main problem of the \textit{constraint reward} ($\mathrm{CoR}$) in MixGAIL \cite{b30} is that this term can only be used on binary classified trajectories or state-action pairs. If each state-action pair's leverage value is not binary, the given data's potential is not fully utilized. Therefore, we propose a \textit{leverage constraint reward} ($\mathrm{LCoR}$), which can fully exploit the given leverage values with varying degrees for training an agent.
The objective function for SSIL can be written as:

\begin{equation}
\small\begin{aligned}
\underset{\pi\in\Pi}{\text{min }}\underset{D}{\text{max}}
& E_\pi[\log(D(s, a))-\eta \mathrm{LCoR}(s, s_T, l_T)] \\ 
& + E_{\pi_E}[\log(1-D(s, a))]-\lambda H(\pi),
\end{aligned}
\label{eq:objective}
\end{equation}
where the notations are the same as that of Section \ref{ssec:MixGAIL}, except for $s_T$ being a set of states of both labeled and unlabeled demonstrations, and $l_T$ being the matching set of leverage value for $s_T$ calculated from Section \ref{ssec:confidence}. LCoR is defined as follows:

\begin{equation}
\begin{aligned}
\mathrm{LCoR}(s,s_T,l_T) &= \frac{\sum_{j=1}^{|l_T|}l_{T_j}
\left(1+\frac{\|s-s_{T_j}\|_1}{\alpha}
\right)
^{-\frac{\alpha+1}{2}}}{\sum_{j=1}^{|l_T|}
\left(
1+\frac{\|s-s_{T_j}\|_1}{\alpha}
\right)
^{-\frac{\alpha+1}{2}}} 
\end{aligned}
\end{equation}
where $|l_T|$ is the size of a leverage value set, $s_{T_j}$ and $l_{T_j}$ are the state value and its corresponding leverage value of the $j$th demonstration, and the value of $\alpha$ is larger than $-1$. 
The $\mathrm{LCoR}$ value adjusts the reward function inside the GAIL algorithm by giving a reward based on leverage values which are weighted by the distance to the state $s$. If states with high leverage values are close, LCoR becomes larger. This allows LCoR to operate as a positive reward incentive, as states with high leverage value are closer to states in the labeled expert demonstrations. On the other hand, if neighboring states have low leverage values, LCoR becomes smaller, causing LCoR to operate as a penalty to the overall reward function. Note that this value only impacts the policy update part, not affecting the discriminator update part.

If $l_{T_j}$ is closer to 1, the corresponding state is closer to expert demonstrations while if $l_{T_j}$ is close to 0, the state is far from expert demonstrations. The modified reward function for the policy is as follows: 

\begin{equation}
r(s, a) = -\log(D(s, a)) + \eta\mathrm{LCoR}(s, s_T, l_T).
\label{eq:target}
\end{equation}

Note that only the labeled state-action pairs are used to calculate the $\mathrm{-\log(D(s,a))}$ in (\ref{eq:target}), and the labeled and unlabeled dataset are used to calculate the value of $\mathrm{LCoR}$.

During training, the discount factor, $\eta$, is decayed. However, each decay causes the target value of the value function to fluctuate. From (\ref{eq:target}), the target value consists of the discriminator value and the $\mathrm{LCoR}$ value. Both values are estimated using neural networks. To mitigate the fluctuation of the value function, $\eta$ is decayed if the difference of the discriminator value function before and after the RL algorithm update is larger than the difference of $\mathrm{LCoR}$ value function before and after the RL algorithm update when both functions are given initial state. The discriminator value function and the $\mathrm{LCoR}$ value function are shown below: 

\begin{eqnarray}
\begin{aligned}
V_\pi(s) &= {V_D}_\pi(s) + \eta {V_C}_\pi(s),\\
{V_D}_\pi(s) &= {E}_{\pi} \left[\sum\limits_{k=t}^T -\gamma^{k-1}\log D(s_k, a_k)\middle|s_t=s\right],\\
{V_C}_\pi(s) &= {E}_{\pi} \left[\sum\limits_{k=t}^T \gamma^{k-1}\mathrm{LCoR}(s_k, s_E, s_N)\middle|s_t=s\right], \\
\Delta {V_C} &= {V_C}_{\pi'}(s_0) - {V_C}_\pi(s_0), \\
\Delta V_D &= {V_D}_\pi'(s_0) - {V_D}_\pi(s_0),
\end{aligned}
\end{eqnarray}
where $V_D$ is the value function of the discriminator, $V_C$ is the value function of $\mathrm{LCoR}$, $s_0$ is an initial state, and $\gamma$ is the decay factor. The $\eta$ is decayed by multiplying $\gamma$ when $\Delta V_D > \Delta V_C$.

\begin{algorithm}[t]
\small
    \caption{SSIL}\label{euclid}
    \begin{algorithmic}[1]
    \REQUIRE Expert demonstrations $\tau_E \sim \pi_E$, unlabeled demonstrations $\tau_{U}  \sim \pi_{U}$, number of policies $n$, expert states set $s_E = \{s_e|\forall{(s_e,a_e)} \in \tau_E\}$, unlabeled states set $s_U = \{s_u|\forall{(s_u,a_u)} \in \tau_U\}$, total state set $s_T = s_E \cup s_U$, initial policy and discriminator parameters $\theta_0,~\omega_0$, weighting parameter of $\mathrm{LCoR}$ $\eta\ge0$, $\eta$'s decay rate $\epsilon$
    \STATE{Train variational autoencoder by expert demonstrations $\tau_E$}
    \STATE{Feed unlabeled demonstrations to trained network and collect reconstruction loss set $U$ for each demonstration}
    \STATE{Normalize the reconstruction loss set $U$ into $l_T$, smallest value to 1 and largest value to 0}
    \FOR{$i= 0, 1, 2, \ldots$}
        \STATE {Sample demonstrations $\tau_i \sim \pi_{\theta_i}$}
        \STATE {Update the discriminator parameters from $\omega_i$ to $\omega_{i+1}$with the gradient
        \begin{equation*}
         \hat{E}_{\tau_i} [\nabla_\omega \log(D_\omega(s, a))] + \hat{E}_{\tau_E} [\nabla_\omega \log(1 - D_\omega(s, a))]
        \end{equation*}}
        \STATE{Take a policy step from $\theta_i$ to $\theta_{i+1}$, using the TRPO rule with cost function $\log(D_{\omega_{i+1}}(s,a))-\eta\mathrm{LCoR}(s,s_T,l_T)$. Specifically, take a KL-constrained natural gradient step with
        \begin{equation*}
         \hat{E}_{\tau_i} [\nabla_\theta \log\pi_\theta (a|s)A_{\pi_{\theta}}(s, a)] - \lambda\nabla_\theta H(\pi_\theta)
        \end{equation*}
        where
        \begin{eqnarray*}
            A_{\pi_{\theta}}(s, a)= 
            -\log D_{\omega_{i+1}}(s, a) + \gamma {V_D}_{\pi_{\theta}}(s')\\
            - {V_D}_{\pi_{\theta}}(s)+ \eta(\mathrm{LCoR}(s,s_T,l_T)\\
            + {V_C}_{\pi_{\theta}}(s') - {V_C}_{\pi_{\theta}}(s)),
        \end{eqnarray*}
        and $s'$ is the next state after state $s$.
        }

        \STATE {update $\eta \gets \epsilon\eta$, when  $\Delta V_D > \Delta {V_C}$
        }
      \ENDFOR
    \end{algorithmic}
\end{algorithm}

During training, we estimate discriminator $D$ and policy $\pi$ using $D_\omega$ and $\pi_\theta$, where $\omega$ and $\theta$ are weights of the neural networks for $D_\omega$ and $\pi_\theta$, respectively. The training process is explained in Algorithm~\ref{euclid}.

%%%%%%%%%%%%%%%%%%%%%%%%%%%%%%%%%%%%%%%%%%%%%%%
%Instead of describing setup for both TORCS and RC car then results for both, explain the setup and results for TORCS and then explain the setup and results for RC car. 
%%%%%%%%%%%%%%%%%%%%%%%%%%%%%%%%%%%%%%%%%%%%%%%
\section{Experimental Results}
\subsection{Simulation Experiment}
We evaluate the proposed algorithm on a race car simulator called TORCS \cite{b12}. In this experiment, a total of 49,200 state-action pairs are collected as demonstrations. The first 1,200 expert state-action pairs are used as the labeled expert demonstrations. The other 48,000 state-action pairs, which consist of 24,000 expert demonstrations and 24,000 negative demonstrations, are used as unlabeled demonstrations. GAIL and MixGAIL are used as baselines for the evaluation of the proposed method. In addition, SSIL using leverage values derived from different algorithms, GPR and MDN, are compared with the proposed algorithm.

\begin{comment}
\begin{figure*}[t!]
\begin{subfigure}{.48\textwidth}
  \centering
  \includegraphics[width=\linewidth]{total_rewards_inonegraph.png} 
\end{subfigure}
\begin{subfigure}{.48\textwidth}
  \centering
  \includegraphics[width=\linewidth]{total_variance_inonegraph.png}  
\end{subfigure}
\caption{Scaled performance and standard deviation of GAIL, MixGAIL, SSIL (MDN), SSIL (VAE), SSIL(GPR), and SSIL(windowVAE) on TORCS. GAIL used 49,200 state-action pairs, which consists of 1,200 labeled expert state-action pairs and 48,000 unlabeled state-action pairs, while MixGAIL and SSIL used 100, 300, 600, 1,200 labeled expert state-action pairs each and 48,000 unlabeled state-action pairs. MixGAIL represents the MixGAIL algorithm trained with an equal number of expert demonstrations to SSIL. The algorithms in the parentheses next to SSIL is the algorithm used to calculate the leverage value. We tested GAIL, MixGAIL, and SSIL three times, each using different random seeds. The left graph shows the average performance over three trials. The right figure is the variance trials scaled to the expert data's performance.}
\label{fig:GAILvsLevMixGAIL}
\end{figure*}
\end{comment}

\subsubsection{Simulation Environment Setup}

\begin{comment}
The TORCS \cite{b12} environment consists of three-dimensional action space and 29-dimensional observation space. The action space is composed of steering, throttle, and brake. The observation space is composed of the velocity vector, 19 range finder sensors, a deviation value from the centerline, the angle between the car and the direction of the track, rotation speeds of wheels, and RPM of the engine. The environment's objective is to drive around the track without colliding with walls or going backward.
\end{comment}

We designed an underlying reward function for the evaluation index of the trained agent's performance. The reward function is shown as below:
\begin{equation}
    R = v_x \cos(\theta)-|v_y \sin(\theta)|- 2 v_x |d \sin(\theta)| - v_y \cos(\theta),
\end{equation}
where $v_x$ and $v_y$ indicates speed of the car along x-axis and y-axis, respectively, $\theta$ indicates the angle between the car direction and the direction of the track axis, which is the center line of the lane, and $d$ is the distance between the car and the track axis.
Note that the underlying reward is only used for evaluation and not used during training.

\subsubsection{Collecting Expert and Negative Demonstrations}
\label{part:datacollection}
The expert demonstrations are collected from a trained PPO\cite{ppo} network. The negative demonstrations are collected from an under-trained PPO network. The unlabeled data of this part consists of part of expert demonstrations and all of the negative demonstrations. A half of the unlabeled data is expert demonstrations, and the other half consists of negative demonstrations. We changed the amount of labeled data to show the efficiency of SSIL. On average, one expert trajectory consists of 2,400 state-action pairs, and one state-action pair is composed of 29 states and two actions.

\subsubsection{Baseline Algorithms for Comparison}
For a fair comparison among VAE, windowed VAE, GPR, and MDN, each evaluation method is tested by using it for training SSIL. For all the compared algorithms, the labeled data are used for training. After training, the unlabeled data are given to each algorithm to obtain leverage values using (\ref{eq:l_cal}). VAE and windowed VAE use the only state for evaluating the unlabeled data. 

On the other hand, GPR and MDN use state-action pairs, where the input to the algorithm is the state and the output of the algorithm is the action. In the case of GPR, the uncertainty estimation of each state-action pair can be obtained by calculating the standard deviation of the predicted output. In the case of MDN, the uncertainty estimation of each state-action pair can be calculated by adding the epistemic uncertainty and aleatoric uncertainty \cite{b32}. In the case of MDN and GPR, the reconstruction error in the equation is replaced with estimated uncertainty.

\subsubsection{Simulation Results}
We tested GAIL, MixGAIL, and SSIL three times, each using different random seeds.
%and plotted a graph of rewards per training step with averages and standard deviations.

\begin {table*}[t]
\begin{center}
\begin{tabular}{ | c | c | c | c | c | c | c | c | c | c |}
 \hline
 labeled state-actions & GAIL & GAIL expert & MixGAIL & SSIL (GPR) & SSIL (MDN) & \textbf{SSIL (VAE)} & \textbf{SSIL (windowVAE)} \\
 \hline

 100 & \textbf{0.564}/0.077 & 0.263/0.057 & 0.049/\textbf{0.002} & 0.051/0.044 & 0.102/0.008 & 0.055/0.017 & 0.325/0.161\\

 300 & 0.564/0.077 & 0.370/0.169 & 0.598/0.125 & 0.514/\textbf{0.021} & \textbf{0.609}/0.085 & 0.605/0.125 & 0.525/0.071\\

 600 & 0.564/0.077 & 0.637/0.129 & 0.586/0.008 & 0.643/0.118 & 0.614/0.102 & \textbf{0.795}/0.054 & 0.763/\textbf{0.002}\\

 1200 & 0.564/0.077 & 0.768/0.095 & 0.886/\textbf{0.003} & 0.880/0.020 & 0.873/0.014 & 0.871/0.013 & \textbf{0.887}/0.022\\
 \hline
\end{tabular}
\caption{The scaled performance (left) and standard deviation (right) of GAIL, MixGAIL, and SSIL on the TORCS domain. The categories written in bold are the proposed algorithm, and the figure of the algorithm that has the best performance and the most negligible standard deviation in each comparison is written in bold also. GAIL expert indicates the GAIL trained with only labeled demonstrations. Each figure shows the performance (left) and standard deviation (right) of each algorithm scaled to the perfect score for each agent.} 
\label{fig:TORCSGAILresult}

%\vspace{-10pt}
\end{center}
\end{table*}

\begin{comment}
\begin {table*}[t]
\begin{center}
\begin{tabular}{ | c | c | c | c | c | c | c | c | c |}
 \hline
 labeled state-actions & GAIL & MixGAIL & SSIL (GPR) & SSIL (MDN) & \textbf{SSIL (VAE)} & \textbf{SSIL (windowVAE)} \\
 \hline
 100 & 0.077 & \textbf{0.002} & 0.044 & 0.008 & 0.017 & 0.161\\

 300 & 0.077 & 0.125 & \textbf{0.021} & 0.085 & 0.125 & 0.071\\

 600 & 0.077 & 0.008 & 0.118 & 0.102 & 0.054 & \textbf{0.002}\\

 1200 & 0.077 & \textbf{0.003} & 0.020 & 0.014 & 0.013 & 0.022\\

 \hline
\end{tabular}
\caption{The scaled standard deviation of GAIL, MixGAIL, and SSIL on the TORCS domain.
The categories written in bold are the proposed algorithm, and the figure with the most negligible standard deviation in each comparison is written in bold also. Each figure shows the standard deviation of each algorithm scaled to the perfect score for each agent.} %write the real figure here
\label{fig:TORCSGAILvariance}
\end{center}
\end{table*}

\end{comment}

The comparative results with SSIL (VAE) and SSIL (windowVAE) are shown in Table~\ref{fig:TORCSGAILresult}. We conduct the experiment based on the number of labeled data's state-action pairs. 100, 300, 600, 1,200 state-action pairs indicates 8.3\%, 25\%, 50\%, 100\% of the labeled expert demonstrations, and 0.20\%, 0.61\%, 1.22\%, 2.44\% of all demonstrations, respectively. 

\paragraph{SSIL vs GAIL}
First, to show the robustness of the proposed algorithm to the quality of data, SSIL is compared with GAIL using labeled demonstrations and unlabeled demonstrations as input to the discriminator. For GAIL, the unlabeled trajectories are given with the expert trajectories to the discriminator. For SSIL, the labeled trajectories are given to the discriminator and $\mathrm{LCoR}$, while the unlabeled data is given to $\mathrm{LCoR}$ only. In addition, the comparison shows how much the labeled data should be given to SSIL in order to outperform GAIL. 

$\mathbf{100~Expert~Demonstrations.}$ 
The results of SSIL trained with leverage values from VAE (SSIL (VAE)) and windowed VAE (SSIL (windowVAE)) show that when the number of input expert state-action pairs is 100, the results are worse than using mixed input of expert and negative trajectories directly to GAIL.
The performance of the experiment on 100 labeled state-action pairs and 48,000 unlabeled state-action pairs is only 9.75\% and 57.6\% of GAIL's performance for SSIL (VAE) and SSIL (windowVAE), respectively, as shown in Table~\ref{fig:TORCSGAILresult}. The result can be interpreted as; VAE sees data that is not seen during training, calculates unseen data's reconstruction high, causing a misallocation of the leverage value and worse training results than GAIL. 

$\mathbf{300~Expert~Demonstrations.}$ 
If 300 labeled state-action pairs and 48,000 unlabeled state-action pairs are used, the SSIL (windowVAE) shows 93.1\% of the performance of GAIL, as shown in Table~\ref{fig:TORCSGAILresult}. On the other hand, the SSIL (VAE) shows an improvement of 7.27\% over GAIL.
The experimental result can be interpreted that in this domain, if 0.61\% of demonstrations are known as expert demonstrations, SSIL (VAE) shows better performance than GAIL. In the case of SSIL (windowVAE), it can be inferred that more data is needed to encode two consecutive states.

$\mathbf{600~\&~1200~Expert~Demonstrations.}$ 
If the number of input data is more than 600, the SSIL (VAE) and SSIL (windowVAE) show higher performance and lower standard deviation than GAIL. In the case of 600 input data, SSIL (VAE) and SSIL (windowVAE) shows 41.0\% and 35.3\% improvements over GAIL, respectively, as shown in Table~\ref{fig:TORCSGAILresult}. In the case of 1200 input data, SSIL (VAE) and SSIL (windowVAE) shows 54.4\% and 57.3\% better performance than GAIL. The comparison with GAIL with mixed quality demonstrations and SSIL shows that rather than using mixed qualities of data as input to GAIL, evaluating each data and exploiting them during training positively affects the performance.

\paragraph{SSIL vs GAIL with only labeled demonstrations}
SSIL (VAE) and SSIL (windowVAE) are compared with GAIL using only labeled demonstrations in the second experiment. In the case of 300, 600, and 1200 expert demonstrations, SSIL (VAE) shows 63.5\%, 24.8\%, and 13.4\% improvements over GAIL, respectively, and SSIL (windowVAE) shows 41.9\%, 19.8\%, and 15.5\% improvement over GAIL, respectively. However, in the case of 100 input data, SSIL (VAE) shows 20.9\% of GAIL's performance, while SSIL (windowVAE) shows 23.5\% of improvement over GAIL. Therefore, it can be inferred that using unlabeled data and exploiting their leverage values during training positively affects the performance. 

\paragraph{SSIL vs MixGAIL}
In the third experiment, SSIL (VAE) and SSIL (windowVAE) are compared with MixGAIL using labeled and unlabeled demonstrations. For MixGAIL, the labeled data are used for $s_E$ in $\mathrm{CoR}$ and discriminator input, and the unlabeled data are used for $s_N$ in $\mathrm{CoR}$, where the unlabeled data are used as negative data. This comparison empirically proves that SSIL has the ability to estimate the quality of each demonstration.

\begin{table}[t]
\begin{tabular}{|l|l|l|l|l|}
\hline
          & 0-25\%  & 25\%-50\% & 50\%-75\% & 100\% (expert) \\ \hline
VAE       & 0.7236 & 0.8474  & 0.8586  & 0.9776        \\ 
WindowVAE & 0.6959 & 0.7333  & 0.7550  & 0.9554        \\ \hline
\end{tabular}
\caption{Leverage value calculated from different qualities of data. The leverage value calculation algorithms are trained with four different qualities of the dataset. 
Each dataset consists of different qualities of data; 0 to 25\% of maximum reward, 25\% to 50\% of maximum reward, 50\% to 75\% of maximum reward, and 100\% of maximum reward (expert data)}

\label{fig:leveragecomparison}
%\vspace{-10pt}
\end{table}

\begin{comment}
\begin{figure}[t]
  \centering
  % include first image
  \includegraphics[width=0.88\linewidth]{total_leverage_value.png} 
\caption{Leverage value calculated from different qualities of data. The leverage value calculation algorithms are trained with four different qualities of the dataset. Each dataset consists of different qualities of data; 0 to 25\% of maximum reward, 25\% to 50\% of maximum reward, 50\% to 75\% of maximum reward, and 100\% of maximum reward (expert data)
}
\label{fig:leveragecomparison}
%\vspace{-10pt}
\end{figure}

\end{comment}

SSIL (VAE) results using 100, 300, and 600 expert state-action pairs as input have 12.2\%, 1.17\%, and 36.8\% improvements over MixGAIL, as shown in Table~\ref{fig:TORCSGAILresult}. In the case of SSIL (windowVAE), results using 100, 600, and 1200 expert state-action pairs as input show improvements of 563\%, 31.2\%, and 0.112\% compared to MixGAIL, as shown in Table~\ref{fig:TORCSGAILresult}.
SSIL (VAE) using 1200 expert state-action pairs and SSIL (windowVAE) using 300 expert state-action pairs show 87.8\% and 98.3\% of the performance of corresponding results of MixGAIL, as shown in Table~\ref{fig:TORCSGAILresult}.
From the results, it can be inferred that SSIL has an advantage over MixGAIL when the expert data is scarce, and the SSIL shows a similar ability to MixGAIL when the data is relatively abundant. In addition, the comparison results show that the discrimination process using VAE is reliable. 

\paragraph{SSIL (VAE), SSIL (windowVAE) vs SSIL (MDN), SSIL (GPR)}
In the fourth experiment, the different algorithms for evaluating the unlabeled data are compared. Leverage value evaluation methods, VAE and windowed VAE, are compared with GPR and MDN.
The comparison is shown by the performance of SSIL trained with the leverage value calculated from each evaluation method. This experiment verifies the validity of the proposed leverage value quality assessments.

The performances of SSIL trained using leverage values from GPR (SSIL (GPR)) and MDN (SSIL (MDN)) were relatively poor compared to SSIL (VAE) and SSIL (windowVAE), except for the case of SSIL (MDN) using 300 expert state-action pairs for training. Especially, SSIL (GPR) showed worse performance than SSIL (windowVAE) in every case. The performance gain of SSIL (windowVAE) over SSIL (GPR) varies from 2.14\% to 537\%. SSIL (MDN) showed 0.661\% to 85.4\% better performance than SSIL (VAE) in most cases except for the case using 600 state-action pairs to input, as shown in Table~\ref{fig:TORCSGAILresult}. However, except for the case using 300 state-action pairs to input, SSIL (windowVAE) shows the superior result to SSIL (MDN), with performance improvement varying from 1.6\% to 219\%, as shown in Table~\ref{fig:TORCSGAILresult}. Considering the standard deviation results shown in Table \ref{fig:TORCSGAILresult}, the best algorithm among the three evaluation methods is SSIL (VAE), as we think that SSIL (VAE) is the sweet spot between standard deviation and performance. Therefore, one can claim that the proposed leverage value quality assessment is adequate for the proposed algorithm.

\paragraph{Leverage Value Comparison}
Finally, the correlations of the leverage value and the quality of demonstrations are shown using different qualities of negative data. The qualities of negative data are calculated based on the reward function of the environment.
%First, we compare the leverage value between expert and negative demonstrations. Second, 
We compare the leverage value of different qualities of negative demonstrations to show that the leverage value reflects the quality of data. %To confirm the second hypothesis, 
The quality of the negative data is split into three parts: 0 to 25\% of the maximum score, 25\% of the maximum score to 50\% of the maximum score, and 50\% of the maximum score to 75\% of the maximum score. The expert data is also used to compare this algorithm for the baseline. This comparison verifies the positive correlation between quality and leverage value.

The results show that the leverage value and the quality of demonstrations have a positive correlation in both VAE and windowed VAE,  as shown in Table~\ref{fig:leveragecomparison}. Both results of the evaluation algorithms show that the evaluation methods (VAE, windowVAE) validly evaluate the reliability of unlabeled demonstrations.

\subsection{Real-World RC Car Experiment}
We apply the proposed method to learn an RC Car controller in the real world to empirically prove that not only the reconstruction loss is an effective method of discriminating and labeling unlabeled data, but also the performance of the proposed method stands out among the compared methods.

\subsubsection{RC Car Environment Setup}
\begin{figure}[t]
    \centering
\includegraphics[width=0.9\linewidth]{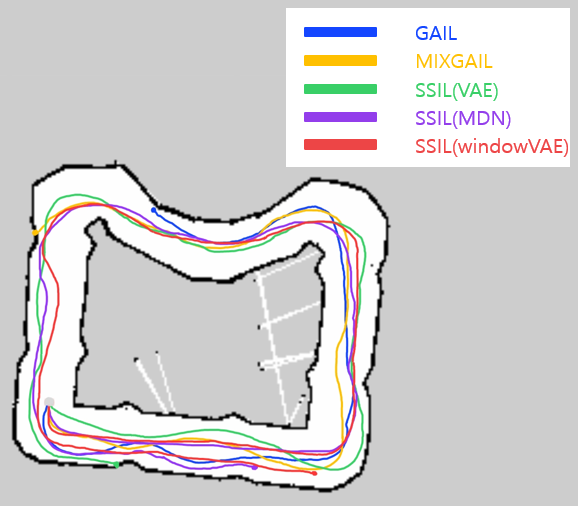}
  \caption{Bird-eye view of the track used for the experiment and its experimental results.}
  \label{birdeyeview}
\end{figure}

The environment's input dimension is 20, which consists of the 19 LiDAR sensor values and the action from a timestep before. The action dimension is one, which represents the steering of the RC car. The initial state's action is given as zero. The objective of this task is to drive around the circuit (see Figure \ref{birdeyeview}). If a collision occurs before rotating around the circuit once, the trajectory is ended immediately.

In this experiment, we do not change the number of expert state-action pairs. One expert trajectory is composed of 401 state-action pairs. We pre-train all five algorithms with behavior cloning\cite{BC} to initialize the training. Similar to the simulation environment, the unlabeled data will be given to MixGAIL as negative demonstrations. Also, the leverage values of input labeled expert data are set to 1.

This experiment's reward function is proportional to how much the RC car has traveled along the track. Since we give the RC car a constant speed, we used the timestep as the performance measure. 

\begin{figure}[t]
    \centering
  \includegraphics[width=0.9\linewidth]{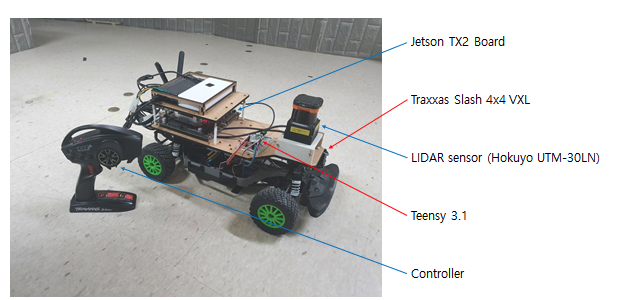}
  \caption{Snapshot of RC car image.}
  \label{fig:rccarpicture}
  %\vspace{-10pt}
\end{figure}

\begin{comment}
\begin{figure}[t]
  \centering
  % include first image
  \includegraphics[width=0.8\linewidth]{Racecar_Lev_final.png}  
  \caption{Scaled performance of GAIL, MixGAIL, and SSIL on RC car. Bold lines are average values over three trials, and the colored area indicates the standard deviation over three trials. The dotted line represents 90\% of the expert score given.}
  \label{fig:rccarresult}
\end{figure}
\end{comment}

\begin {table*}[t]
\begin{center}
\begin{tabular}{ | c | c | c | c | c | c | }
 \hline
 evaluation metric & GAIL & MixGAIL & SSIL (MDN) & \textbf{SSIL (VAE)} & \textbf{SSIL (windowVAE)}\\
 \hline
 maximum score & 0.679 & 0.721 & \textbf{0.813} & 0.786 & 0.808 \\ %exact numbers

 standard deviation & 0.086 & 0.051 & 0.089 & \textbf{0.050} & 0.051\\
 \hline
\end{tabular}
\caption{The scaled performance of GAIL, MixGAIL and SSIL(MDN, VAE, windowed VAE) on RC car domain. The expert score of the environment is 400, and each algorithm's scores are normalized by provided expert score.}
\label{fig:RCcarnum}
\end{center}

%\vspace{-20pt}
\end{table*}

\subsubsection{Collecting Expert and Negative Demonstrations}
Expert demonstrations are collected by controlling the RC car manually. The expert demonstrations consist of 4,010 state-action pairs, which are ten trajectories. We use 2,005 state-action pairs, which is a half of the expert demonstrations, as labeled state-action pairs. The other 2,005 state-action pairs are given as unlabeled data with other 1,170 negative demonstrations by selecting 250-300 state-action pairs each from four different qualities of under-trained policies. The negative demonstrations were collected from under-trained GAIL. These demonstrations include episodes where RC car collides with walls. Note that to show the robustness of our algorithms in the real world, the track used to collect data is different from the track used to train the algorithm.

\subsubsection{RC Car Experimental Results}
GAIL, MixGAIL, and SSIL are trained three times, each using different random seeds. The plotted graph is the average rewards per training step and standard deviations. The snapshots of the experiments are shown in Figure \ref{fig:rccarpicture}.

The numerical results are shown in Table~\ref{fig:RCcarnum}. The results of SSIL with leverage values calculated from VAE are 9.02\% higher than MixGAIL and 15.8\% higher than the original GAIL, but its average performance was worse than that of SSIL trained with MDN leverage values. As shown in Table \ref{fig:RCcarnum}, the SSIL using VAE has a lower average standard deviation than GAIL and SSIL trained with MDN leverage values and shows a similar average standard deviation to MixGAIL. 

The windowed VAE version of SSIL also shows higher performance than GAIL and MixGAIL and similar performance to SSIL trained with MDN leverage value. The windowed VAE version of SSIL shows performance gains of 19\% and 12.1\% compared to GAIL and MixGAIL, respectively, and has a lower standard deviation than GAIL and SSIL trained with MDN leverage values, as shown in Table \ref{fig:RCcarnum}.  

We would like to emphasize that the standard deviation of GAIL is the second-highest since the algorithm is tested on a similar but different track from the track used to collect the training data. MixGAIL showed worse performance than all SSIL methods because the unlabeled data were not labeled correctly, causing it to be seen as negative, although it is actually expert data. This process causes the \textit{constraint reward} to be contaminated with wrong information. On the other hand, SSIL trained with VAE and windowed VAE safely discriminates and labels expert data, causing the standard deviation during training to lower and show better performance. This labeling process allows the proposed algorithm to gain valid information about each unlabeled data from the \textit{leverage constraint reward}, allowing our agent to explore the environment safely. Also, SSIL trained with MDN shows a similar maximum score to windowed VAE but has a much higher standard deviation, from which we can infer that MDN leverage value is less stable than windowed VAE value.

\section{Conclusion}
This paper has considered handing unlabeled, perhaps sub-optimal, demonstrations in imitation learning. We have proposed a method to discriminate the demonstration at the state-action pair level. Moreover, we have proposed a new reward incentive called \textit{leverage constraint reward} that is suited for continuous evaluation of demonstrations, which is called leverage. As shown in Table~\ref{fig:leveragecomparison}, the estimated leverage values correspond to the quality of data.
%TORCS 에서 잘 나온다??
The SSIL has been able to show better performance than the baseline algorithms in TORCS simulations, when the labeled expert demonstrations are less than 1\% of the total demonstrations.
%The SSIL has outperformed GAIL and MixGAIL when only more than 0.61\% of demonstrations are labeled expert demonstrations.
In the RC car experiment, SSIL has shown a significant performance improvement over GAIL and MixGAIL in terms of the traveral distance when given similar amounts of data. 

\bibliographystyle{IEEEtran}
\bibliography{myBIB}

\end{document}